\documentclass{article}

     \PassOptionsToPackage{numbers, compress}{natbib}

\usepackage[preprint]{neurips_2020}

\usepackage[utf8]{inputenc} 
\usepackage[T1]{fontenc}    
\usepackage{hyperref}       
\usepackage{url}            
\usepackage{booktabs}       
\usepackage{amsfonts}       
\usepackage{nicefrac}       
\usepackage{microtype}      
\usepackage{amsmath}
\usepackage{enumitem}
\usepackage{chemfig}
\usepackage{subcaption}
\usepackage[normalem]{ulem} 
\usepackage[bottom]{footmisc} 

\DeclareMathOperator*{\argmax}{arg\,max}

\title{Graph-Aware Transformer: Is Attention All Graphs Need?}

\author{
  Sanghyun Yoo\qquad Young-Seok Kim\qquad Kang Hyun Lee\qquad Kuhwan Jeong\AND
  Junhwi Choi\qquad Hoshik Lee\qquad Young Sang Choi\\\\
  Samsung Advanced Institute of Technology\\
  \texttt{\{sam.yoo,ys24.kim,kh1216.lee,kuhwan.jeong,junhwi.choi,hoshik.lee,macho\}}\\
  \texttt{@samsung.com} \\
}

\begin{document}

\maketitle

\begin{abstract}

Graphs are the natural data structure to represent relational and structural information in many domains. 
To cover the broad range of graph-data applications including graph classification as well as graph generation, it is desirable to have a general and flexible model consisting of an encoder and a decoder that can handle graph data. Although the representative encoder-decoder model, Transformer, shows superior performance in various tasks especially of natural language processing, it is not immediately available for graphs due to their non-sequential characteristics. To tackle this incompatibility, we propose GRaph-Aware Transformer (GRAT), the first Transformer-based model which can encode and decode whole graphs in end-to-end fashion. GRAT is featured with a self-attention mechanism adaptive to the edge information and an auto-regressive decoding mechanism based on the two-path approach consisting of sub-graph encoding path and node-and-edge generation path for each decoding step. We empirically evaluated GRAT on multiple setups including encoder-based tasks such as molecule property predictions on QM9 datasets and encoder-decoder-based tasks such as molecule graph generation in the organic molecule synthesis domain. GRAT has shown very promising results including state-of-the-art performance on 4 regression tasks in QM9 benchmark. 

\end{abstract}







\section{Introduction}

The ability to deal with graphs has many applications~\cite{Zhou2018corr}: designing efficient chip architectures~\cite{pmlr-v97-zhang19e}, predicting missing information in knowledge bases~\cite{hamaguchi2017knowledge}, rating items in social recommendation systems~\cite{wenqi2019www}, predicting molecular properties and generating new molecular structures, which could expedite the discovery process of new drugs and materials~\cite{butler2018ml}.

The crux of the ability to deal with graph data for such applications lies in two parts. The first part is how to create a representation which can effectively capture the relationship between nodes and edges for a given graph. The second part is how to generate a new graph, i.e., a set of nodes and edges, 
satisfying certain requirements. Generally, the representation for a given graph is created by an encoder, and the new graph generation is conducted by a decoder. Therefore, in order to cover the needs of the aforementioned broad application domains, it is desirable to have a general model consisting of the encoder and the decoder, instead of having 
application-specific models. 

Transformer~\cite{vaswani2017} 
has shown state-of-the-art performance in many applications such as automatic speech recognition, neural machine translation, named entity recognition, sentiment analysis, question answering, visual question answering, and more. These applications deal with different kinds of modalities from voice wave and text to image and even multi-modal data. There are many aspects of Transformer, which resulted in such superior performance. Among them, the multi-head self-attention mechanism is arguably the most important one. Considering the power of Transformer and its general applicability to many types of applications dealing with different kinds of modalities, it is a very natural step to apply Transformer to graph-data applications and evaluate its effectiveness. Moreover, since Transformer consists of the encoder 
and the decoder, 
it can serve as a general model which may cover the broad range of the graph-data applications.

The luxury of Transformer, however, is not immediately available for the graph data due to the non-sequential characteristic. To tackle this incompatibility, we consider nodes as a sequence of tokens, one token for each node, where the order of the tokens could be imposed depending on the applications' necessity. 
Edges are reflected when the self-attention weights are computed, since the attention weights are used to decide how much information should be taken from the other tokens (i.e., nodes). With this manipulation, graph data can go through Transformer without losing any meaningful information. Moreover, in order to generate a new graph through the decoder, we add an ability to generate a new node with its associated edges 
to the nodes generated so far
for each decoding step in an auto-regressive manner. 

Our paper's main contributions are as follows:
\begin{itemize}
\item We propose a Transformer-based graph neural network, called Graph-Aware Transformer (GRAT). To the best of our knowledge, GRAT is the first Transformer-based model which can take graphs as input and generate whole graphs as output in end-to-end fashion. Also, since we do not impose any application or domain specific assumptions on GRAT, it is very general and flexible enough to be applicable to any graph-data applications.
\item With GRAT, we study the effectiveness of applying Transformer to graphs in multiple setups. First, we apply GRAT to graph property prediction tasks, where only the encoder part is used to see whether 
the modified self-attention mechanism is able to effectively capture the graph structure. 
Second, GRAT is employed for a task of generating new graphs for input graphs, where, of course, both encoder and decoder parts are used together.
\item Our proposed GRAT model has shown very promising results including state-of-the-art performance on graph property prediction tasks. Based on the results, we would like to argue that a Transformer-based model, i.e., a general encoder-decoder model armed with multi-head self-attention, is a compelling and viable option for graph-data applications. 
\end{itemize}

\section{Transformer}
Since our architecture is based on Transformer \cite{vaswani2017}, we briefly introduce it first. It has achieved state-of-the-art performance for various sequence-to-sequence tasks such as machine translation, which could be attributed to the general encoder-decoder architecture armed with the attention mechanism.

One of the most important features of Transformer is the attention function called \textit{Scaled Dot-Product Attention}, which maps a query and a set of key-value pairs to output as shown below:
\begin{gather}
Attention(Q, K, V) = softmax \left( \frac{QK^{T}}{\sqrt{d_k}} \right) V
\end{gather}\\
where $Q$, $K$, and $V$ are matrices of the same size, and $d_k$ is the dimension of the keys.

Transformer, however, is not immediately available for graph applications in the following reasons. First, even if we can easily map nodes of a graph to a sequence of tokens, Transformer does not provide a way to explicitly take the relational information (i.e., edge information) between nodes. Although the attention mechanism is designed to figure out the implicit relationship between input tokens (i.e., nodes), it may not fully utilize the explicitly given edge information of the input graph. Second, the decoder of Transformer assumes every token highly depends on the very previous token, which is realized by inputting the very previous token as a query when predicting the next token. However, this assumption may not be valid for graphs since it could be better to take into account the overall structure of the generated sub-graph in each decoding step. We describe how GRAT tackles these problems in the next section.

\section{Graph-Aware Transformer}

\subsection{Graph Encoding}
\label{section:graph_encoding}

We carve out nodes and edges of a given graph to fit into Transformer model. Nodes are considered as tokens, one token for each node, where the order of the tokens could be 
imposed depending on the applications' necessity. Edges are reflected when the self-attention weights are computed. More specifically, we allow Transformer to learn the importance of each connection between nodes by reflecting edge information into the scaled dot-product attention.

\citet{maziarka2020molecule} have recently proposed the similar idea which exploits the distance between nodes and the adjacency matrix as well as the original attention values. However, they just use the weighted sum of these three attention values, where the weights are hyper parameters 
which are handcrafted and may vary for each application.

To make our model learn the importance of each connection solely from the data, we use feature-wise transformation \cite{dumoulin2018}. In this method, the attention values are transformed with the scaling factor $\gamma$ and the biasing factor $\beta$, which are generated based on the edge type as the conditioning information.

The attention values, $Attention(Q, K, V)$, are calculated as follows:
\begin{gather}
(\gamma_{ij}, \beta_{ij}) = f_a(e_{ij})\\
Attention(Q, K, V) = softmax \left( \frac{\Gamma \odot (QK^{T}) + B}{\sqrt{d_k}} \right) V
\end{gather}

where $e_{ij}$ indicates the edge type (in one-hot representation) between nodes $i$ and $j$, and $f_a$ is multi-layer perceptrons (MLPs). $\Gamma$ and $B$ are the matrices whose elements at $(i, j)$ position are $\gamma_{ij}$ and $\beta_{ij}$, respectively. The operator $\odot$ indicates the element-wise multiplication.

Note that the way that we reflect the edge type into the attention mechanism is flexible enough to incorporate any additional per-edge features as input to $f_a$. For example, when the distance between nodes $i$ and $j$ is available as an additional edge feature, 
the concatenation of the distance and edge type $e_{ij}$ can be used as input to $f_a$.
Also, depending on applications' characteristics, it might be better to prevent a node from attending on the non-neighbor nodes. This could be achieved by setting corresponding logit values to the negative infinity ($-\infty$) before the softmax function so that the attention values are 
set to zeros. Lastly, when the property of permutation invariance is required, we may simply remove the positional encoding after the input embedding \cite{vaswani2017}.



\subsection{Auto-regressive Graph Decoding}
GRAT's decoder generates a node and its associated edges for each decoding step in an auto-regressive manner. 
The decoding process at time step $i$ consists of two paths: 1) encoding the sub-graph generated by step $i$-$1$ and 2) generating a new node and its associated edges 
to the existing ones. This is the point mainly different from Transformer decoder, since it does not distinguish these two paths. The rationale behind this idea is that generating a new node should not highly depend on the immediate previous node. Instead, due to the relational nature of the graph, it should take into account all the previously generated nodes based on their relationship, i.e, edge information. To reflect this rationale into the decoding step explicitly, we avoid injecting the immediate previous node as input token to the next step directly by introducing the two-path decoding step.

\begin{figure}[ht!]
    \centering
    \begin{subfigure}[c]{0.72\textwidth}
        \centering
        \includegraphics[width=\textwidth]{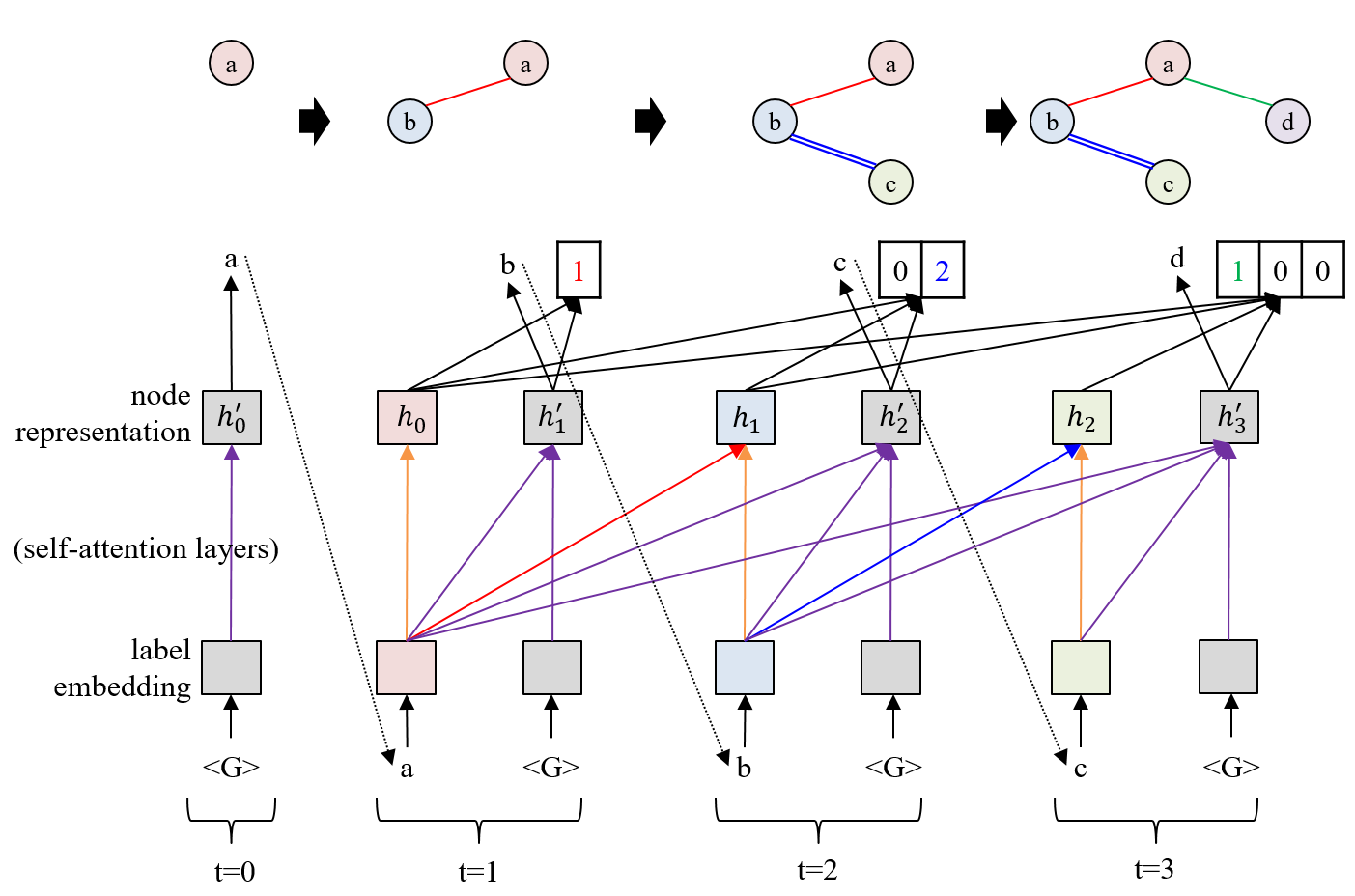}
    \end{subfigure}
    \begin{subfigure}[c]{0.26\textwidth}
        \centering
        \includegraphics[width=\textwidth]{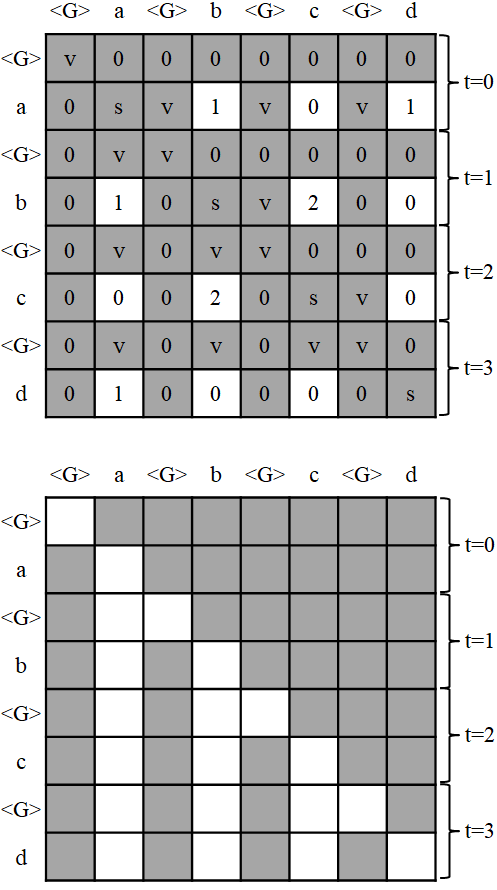}
    \end{subfigure}
    \caption{An example of the two-path decoding steps (left), the corresponding edge matrix (top-right), and the masking matrix (bottom-right)}
    \label{fig:autoregressive_decoding}
\end{figure}

For clarity, we explain the detailed behavior of the two-path decoding step with an example illustrated in Figure~\ref{fig:autoregressive_decoding}. We omit the encoder and the stacked self-attention layers for brevity. In the example, the decoder should generate a graph consisting of 4 nodes with associated edges, where $a$, $b$, $c$, and $d$ are node labels, and the line types of the edges represent edge types. Suppose that the model has already generated the nodes, $a$, $b$, and $c$ with their edges until time step $t$=$2$. Now, according to the two-path decoding step, the first path (i.e., the encoding path) encodes the sub-graph and outputs the nodes' representation. Since the representations for $a$ and $b$, $h_0$ and $h_1$, are already made at $t$=$1$ and $t$=$2$ and can be reused, only the representation $h_2$ for node $c$ is computed at this time by reflecting the representations of all neighbor nodes, only $b$ here though, through the attention mechanism as described in Section~\ref{section:graph_encoding}.

Next, the second path (i.e., node-and-edge generation path) inputs a special token <G> 
and outputs its node representation $h'_{3}$ taking into account all previously generated nodes through the attention again\footnote{Note that the node representations generated on the encoding path and the generation path at time step $i$ are denoted as $h_{i-1}$ and $h'_{i}$, respectively. Also, note that $h_{i-1}$ is reused until the end of the decoding, whereas $h'_{i}$ is used only for the current step and discarded.}, and then generates a label $d$ of a new node. Plus, the associated edges are generated, where the order of the edge type values in the figure corresponds to the order of their node generations. Since the node $d$ is connected to the node $a$ only, the predicted value representing the edge type between node $d$ and $a$ is one, and the others are zeros, each of which means no edge. The decoding process ends if the predicted label is another special token <EOG>.

More formally, if the token <G> is injected at time step $i$, i.e., on the generation path, the representation for a new node, $h'_{i}$ is generated. Then, the label for this $i$-th node, $l_{i}$, and the type of the edge between the $i$-th and $j$-th node, $e_{ij}$ (where $i > j$), can be computed as:
\begin{gather}
l_{i} = \argmax(f_l(h'_{i}))\\
e_{ij} = \argmax(f_e(f_p(h'_{i}), f_p(h_{j})))
\end{gather}

where $h_{j}$ is the representation of the $j$-th node, $f_l$ and $f_e$ are MLPs, and (., .) denotes concatenation. We use an additional MLP, $f_p$, to reduce the dimension of the representation of each node.

The edge matrix used in this example is shown as the top-right matrix in Figure~\ref{fig:autoregressive_decoding}. There are two special edge types, virtual edge ($v$) and self-edge ($s$). The virtual edge $v$, corresponding to the purple line in the figure, represents the edge type unknown yet and is used to attend on existing nodes when generating a new node. The self-edge $s$ (the orange line) indicates the attention to the node itself, which should be distinguished from other edge types. Also, values at the white cells in the matrix represent edge types to be predicted as the decoding proceeds while values at the gray cells are simply given at the beginning of the decoding.

The nodes to be generated after the current time step and the special <G> tokens before the current step should not be visible from each decoding step's perspective. To avoid such invalid reference, we use a masking matrix as shown in the bottom-right matrix in Figure~\ref{fig:autoregressive_decoding}. The upper triangle in this matrix corresponds to the future context as in \citet{vaswani2017}, and the remaining dark cells are for blocking the reference to the previous <G> tokens. Thanks to this masking matrix, all the decoding steps happen in parallel during the training time even though we have explained the decoding steps as a sequential procedure one node at a step.

\subsection{Pretraining Strategies}
\label{section:pretraining_strategies}
Since GRAT resembles Transformer, we can leverage the BERT-style pretraining tasks that enable Transformer architecture to achieve state-of-the-art in many NLP problems~\cite{devlin2019naacl}. The first task used by BERT is the masked LM by which the model is trained to predict the masked words. We tweak this task to mask node labels and their associated edge types for a given graph such that GRAT is trained to predict not only the node labels but also the edge types. The second task is the next sentence prediction task targeted for coming up with a sentence-level representation in contrast to the word-level representation in the masked LM task. We also tweak this task to come up with a graph-level representation such that GRAT is trained to predict graph-level properties (if available) for a given graph. To do so, we introduce another special token <CLS> in front of the input node sequence of the encoder. The token's final hidden vector generated by the encoder is fed into a fully-connected layer, and the subsequent output represents a predicted graph-level property. This type of graph pretraining strategy is inspired by the one used in~\cite{hu2020pretrain}. Both node-level and graph-level pretraining tasks help GRAT have an ability to figure out the relational information between nodes and edges in two different levels.


\section{Experimental Results}
To examine the effectiveness of applying Transformer for graph data and its generality and flexibility, we empirically evaluate GRAT on multiple setups including encoder-based tasks such as graph property predictions and encoder-decoder-based tasks such as graph-to-graph translations.

\subsection{Property Prediction}
\label{section:property_prediction}
For graph property predictions, we use the QM9 benchmark \cite{zhenqin2018rsc}. QM9 is a dataset that provides geometric, energetic, electronic and thermodynamic properties of roughly 130K molecules with up to nine heavy atoms, yielding 12 regression tasks. All molecules are modeled using density functional theory. Each task should predict a property for a given molecule, i.e., a graph consisting of atoms and bonds, i.e., nodes and edges, respectively.  We randomly chose 10000 samples for validation, 10000 samples for testing, and used the rest for training as in \cite{gilmer2017neural}. There are two modes to train models for the tasks: a multi-task model, i.e., one model for all 12 regression tasks, and 12 single-task models, i.e., one model for each regression task.

For the benchmark, we use the encoder of GRAT as our proposed model and compare with two other state-of-the-art models, MPNN \cite{gilmer2017neural} and DimeNet~\cite{klicpera2020directional}.
MPNN is a variant of models belonging to Message Passing Neural Network framework, where all such models are characterized as a common composition, i.e., a message function, an update function, and a readout function. Further details are described in Section~\ref{section:related_work}. DimeNet also belongs to this framework, but it takes into account molecule-specific information such as angles formed between neighbor nodes.

\begin{table}[h]
  \caption{MAE of 12 regression tasks in QM9}
  \label{exp_property}
  \centering
  \begin{tabular}{c|c||ll|lll}
    \toprule
    \multicolumn{2}{c||}{Mode}   & \multicolumn{2}{c|}{Multi-task}        & \multicolumn{3}{c}{Single-task} \\
    \midrule
    \textbf{Target} & Unit   & DimeNet & GRAT & MPNN  & DimeNet & GRAT \\
    \midrule
    \textbf{mu} & D & 0.0775   & \textbf{0.04859} & 0.03 & \textbf{0.0286}   & 0.03898 \\
    \textbf{alpha} & $a_0^3$   & \textbf{0.0649}  & 0.10258  & 0.092& \textbf{0.0469}  & 0.07219  \\
    \textbf{HOMO}  & eV & 0.0451    & \textbf{0.02648} & 0.04257    & 0.0278   & \textbf{0.02228} \\
    \textbf{LUMO}  & eV & 0.0411    & \textbf{0.02783}   & 0.03741    & \textbf{0.0197}   & 0.02053 \\
    \textbf{gap}   & eV & 0.0592   & \textbf{0.03864}   & 0.0688 & \textbf{0.0348}  & 0.035 \\
    \textbf{R2}  & $a_0^2$   & \textbf{0.345}   & 2.68846   & \textbf{0.18}     & 0.331  & 0.76681 \\
    \textbf{ZPVE} &  eV & \textbf{0.00287}  & 0.007    & 0.001524 & \textbf{0.00129}  & 0.00208 \\
    \textbf{U0} & eV & \textbf{0.0129}   & 0.05864   & 0.01935   & \textbf{0.00802}  & 0.0705 \\
    \textbf{U} &  eV & \textbf{0.013}  & 0.05859    & 0.01935 & \textbf{0.00789}  & 0.02825 \\
    \textbf{H} &  eV & \textbf{0.013}   & 0.06014   & 0.01677 & \textbf{0.00811 }  & 0.02549 \\
    \textbf{G} &  eV & \textbf{0.0139}  & 0.05769   & 0.01892  & \textbf{0.00898}  & 0.02505 \\
    \textbf{Cv}  & $cal/molK$ & \textbf{0.0309}   & 0.05512   & 0.04     & \textbf{0.0249}  & 0.03261 \\
    \midrule
    \textbf{stdMAE}  & \% & 1.92   &\textbf{ 1.62}    & 1.7     & \textbf{1.05}   & 1.19 \\
    \textbf{logMAE}  & - & \textbf{-5.07}  & -4.39    & -5.09     & \textbf{-5.57}  & -4.92 \\
    \bottomrule
  \end{tabular}
\end{table}

The mean absolute errors (MAE) of 12 regression tasks in QM9 are shown in Table~\ref{exp_property}. Numbers in bold font represent the state-of-the-art performance for those tasks. For the multi-task mode, GRAT shows state-of-the-art performance on 4 out of 12 tasks. Notably, in terms of the mean standardized MAE (stdMAE)~\cite{klicpera2020directional}, GRAT achieves new state-of-the-art performance across 12 tasks, overall. The corresponding results of MPNN are not available. For the single-task mode, DimeNet shows state-of-the-art performance on 10 out of 12 tasks.
\paragraph{The effect of pre-training}
The first pre-training curriculum consists of two steps: pre-training with QM9 dataset followed by fine-tuning with QM9 dataset. The second one consists of three steps: pre-training with GDB-17 dataset~\cite{Ruddigkeit2012}, then the same two steps of the first curriculum. The GDB-17 dataset consists of 50M molecules, but, for our pre-training, we remove molecules including any atom that does not appear in QM9 dataset, which resulted in around 40M molecules. 
The graph-level tasks for GDB-17 dataset should predict 11 molecule descriptors\footnote{See appendix~\ref{section:add_detail_4.1} for more details of the 11 molecule descriptors.} for a given molecule. 
Since those descriptors are different from the properties of the 12 tasks in QM9 and can be simply obtained by RDKit \cite{rdkit}, 
we clarify that the number of train data points with respect to the ground truth remains the same.

The results of the pre-trainings are shown in Table~\ref{exp_property_pretrain}, where the models pre-trained with the first curriculum and with the second curriculum are denoted as p-GRAT and pp-GRAT, respectively. For the single-task mode, pp-GRAT achieves new state-of-the-art performance on 4 out of 12 tasks. Also, in terms of the stdMAE, pp-GRAT obtains new state-of-the-art performance across 12 tasks, overall. These results are not achievable without the pre-trainings. In addition, the underlined numbers under p-GRAT model represent superior performance than DimeNet even though they are inferior than pp-GRAT.  Lastly, by comparing the results of GRAT with p-GRAT and pp-GRAT, we can see the positive effect of the pre-trainings very clearly throughout all 12 tasks. 



\begin{table} [!b]
  \caption{MAE of 12 regression tasks in QM9 - the effect of pre-training}
  \label{exp_property_pretrain}
  \centering
  \begin{tabular}{c||lll|llll}
    \toprule
    \textbf{Mode}   & \multicolumn{3}{c|}{Multi-task}        & \multicolumn{4}{c}{Single-task} \\
    \midrule
    \textbf{Target}   & DimeNet & p-GRAT  & pp-GRAT & MPNN  & DimeNet & p-GRAT & pp-GRAT\\
    \midrule
    \textbf{mu}       & 0.0775   & \underline{0.03986} & \textbf{0.03461}   & 0.03 & 0.0286   & 0.03022  & \textbf{0.02807}      \\
    \textbf{alpha}   & \textbf{0.0649}  & 0.09815    & 0.08359   & 0.092& \textbf{0.0469}  & 0.05839    & 0.05451      \\
    \textbf{HOMO}  & 0.0451    & \underline{0.02476} & \textbf{0.02175}    & 0.04257    & 0.0278   & \underline{0.02032}  & \textbf{0.01908}      \\
    \textbf{LUMO}  & 0.0411    & \underline{0.02695} & \textbf{0.02231}   & 0.03741    & 0.0197   & \underline{0.01844}  & \textbf{0.01771}      \\
    \textbf{gap}    & 0.0592   & \underline{0.0347}    & \textbf{0.03164}   & 0.0688 & 0.0348  & \underline{0.03134}    & \textbf{0.0297}      \\
    \textbf{R2}  & \textbf{0.345}   & 2.12725   & 1.86826   & \textbf{0.18}     & 0.331  & 0.53468     & 0.63276      \\
    \textbf{ZPVE}& \textbf{0.00287}  & 0.00419  & 0.00534    & 0.001524 & \textbf{0.00129}  & 0.00189 & 0.002      \\
    \textbf{U0} & \textbf{0.0129}   & 0.0433  & 0.04443   & 0.01935   & \textbf{0.00802}  & 0.02629    & 0.02199      \\
    \textbf{U} & \textbf{0.013}  & 0.04419   & 0.05684    & 0.01935 & \textbf{0.00789} & 0.02795  & 0.02511      \\
    \textbf{H} & \textbf{0.013}   & 0.04551   & 0.05639   & 0.01677 & \textbf{0.00811 }  & 0.02734 & 0.02115      \\
    \textbf{G} & \textbf{0.0139}  & 0.04289   & 0.05692   & 0.01892  & \textbf{0.00898}  & 0.0268  & 0.01855      \\
    \textbf{Cv}  & \textbf{0.0309}   & 0.0453  & 0.05403   & 0.04     & \textbf{0.0249}  & 0.02929    & 0.02931      \\
    \midrule
    \textbf{stdMAE}  & 1.92   & \underline{1.4}  & \textbf{1.32}   & 1.7     & 1.05  & \underline{1.01}   & \textbf{0.95}      \\
    \textbf{logMAE}  & \textbf{-5.07}  & -4.60   & -4.57   & -5.09     & \textbf{-.5.57}  & -5.11    & -5.18      \\
    \bottomrule
  \end{tabular}
\end{table}

\subsection{Chemical Reaction Outcomes Prediction}
\label{section:chemical_reaction_outcomes_prediction}
For evaluating the ability of the graph-to-graph translation, we apply GRAT to predict the products of organic chemical reactions given their reactants and reagents using USPTO dataset~\cite{lowe2014} which is filtered and split by~\citet{jin2017}. Although many graph-based solutions have been proposed for the task~\cite{jin2017, do2019, coley2019}, GRAT is the first work that tries to tackle such a task as a graph-to-graph translation in end-to-end fashion, where each data point consists of reactants and reagents as input graphs and products as output graphs.


Table~\ref{exp_comp_table} shows the performance of GRAT and other compared models. MT~\cite{schwaller2019} showing the best performance represents the input and output chemical compounds in SMILES and takes the sequence-to-sequence approach with Transformer. Note that we only take atom labels for nodes and bond types and distances for edges from the graph structure, hence there is loss of information during the transformation of SMILES into the graph structure. We expect that the performance could be improved if GRAT considers the other chemical information carried in SMILES such as formal charges and valences. Adding more complete molecule information to GRAT is remained as future work.

\begin{figure}[t!]
    \centering
    \begin{subfigure}[c]{0.59\textwidth}
        \centering
        \includegraphics[width=\textwidth]{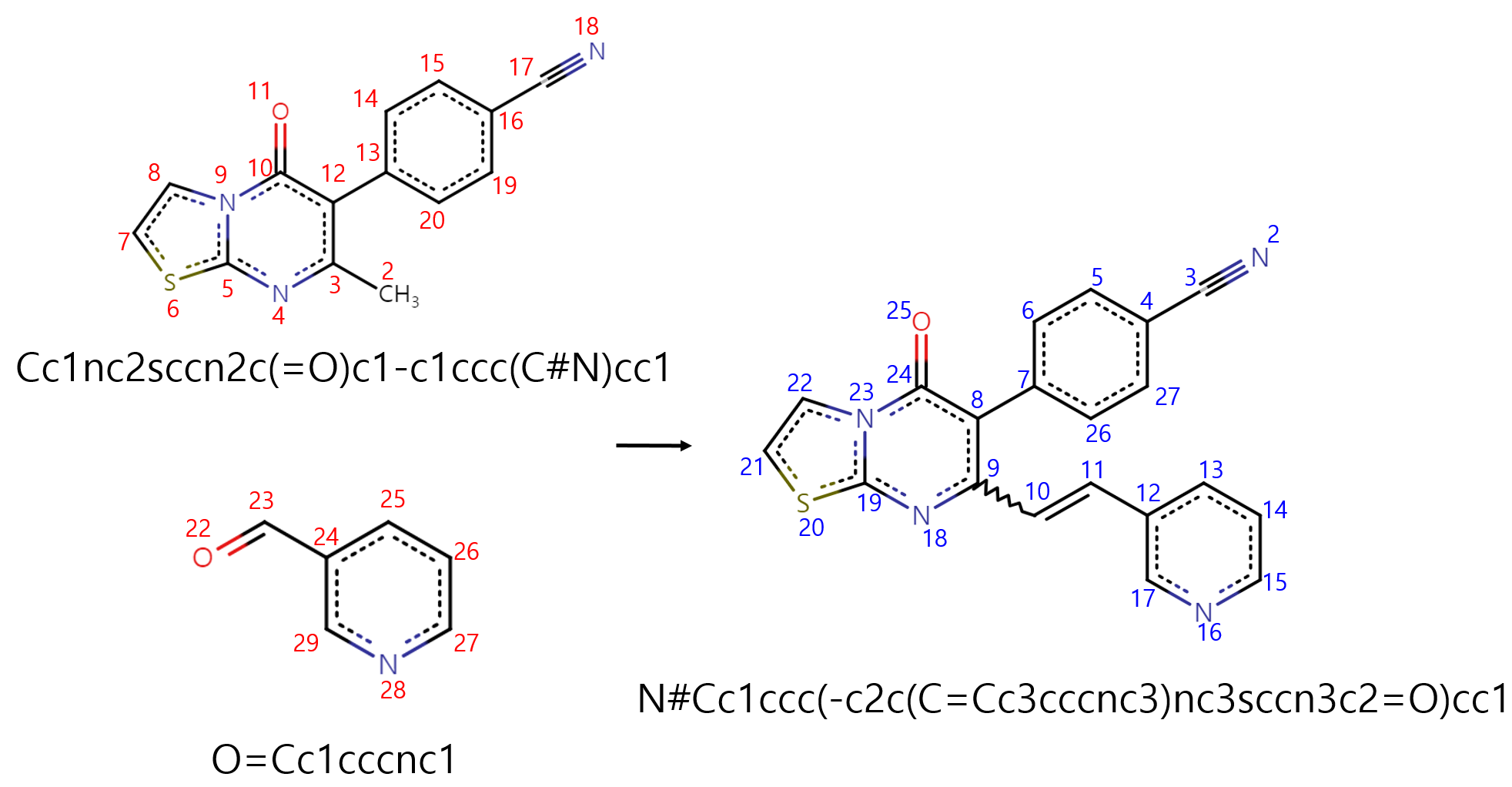}
        \label{fig:reaction_example}
    \end{subfigure}
    \begin{subfigure}[c]{0.4\textwidth}
        \centering
        \includegraphics[width=\textwidth]{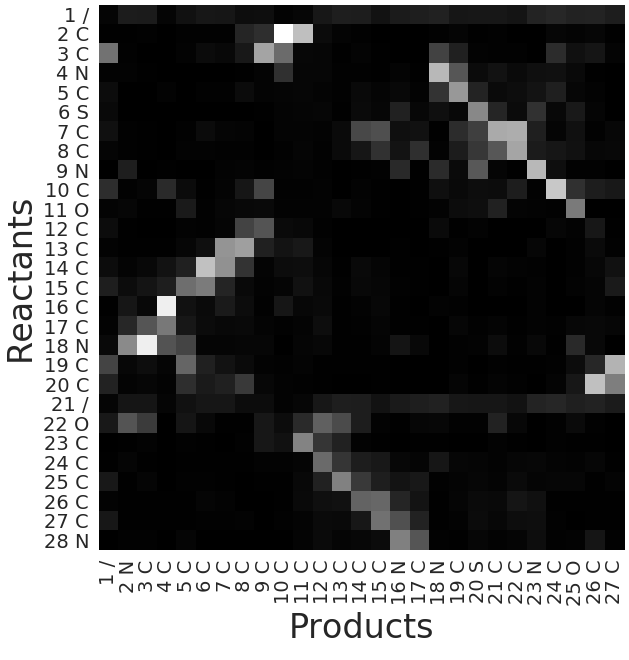}
        \label{fig:attention_weights}
    \end{subfigure}
    \caption{An example of a chemical reaction (left) and visualization of attention values (right)}
    \label{fig:reaction_example_and_attention_weights}
\end{figure}

GRAT has a merit of providing an easier way to explain the reaction mechanism than the SMILES-based model by using the graph structure and the attention values. For example, we can visualize which parts of the reactants are combined to produce a product by pinpointing the participating atoms as shown in Figure~\ref{fig:reaction_example_and_attention_weights},
where two reactants produce a product. (We omit the reagent information for brevity.) The node 2 is combined with 23 in the reactant side (numbers in red) and becomes 10 in the product side (numbers in blue). In the attention matrix (the right side of the figure), we can see that at the generation step for the node 10 in the product side (along the x-axis), the attention values corresponding to the node 2, 22, and 23 in the reactant side (along the y-axis) are higher/brighter than others.



\begin{table}
  \caption{Comparison of Top-1 Accuracy Obtained by the Different Single-Model Methods on the Current Benchmark}
  \label{exp_comp_table}
  \centering
      \begin{tabular}{lllllll}
        \toprule
        \textbf{Models}   & S2S \cite{schwaller2018}  & WLDN \cite{jin2017} & GTPN \cite{do2019} & WLDN5 \cite{coley2019} & MT \cite{schwaller2019}  & GRAT  \\ 
        \midrule
        \textbf{Accuracy} & 80.3 & 79.6 & 82.4 & 85.6  & \textbf{88.8} (90.4\footnotemark[3]) & 88.25      \\
        \bottomrule
      \end{tabular}
\end{table}
\footnotetext[3]{From the model that used the training data augmentation and averaged checkpoints.}

\section{Related Work}
\label{section:related_work}
Graph neural networks (GNNs) are designed for dealing with graph data structures. 
As abstracted by~\citet{gilmer2017neural}, many GNN models can be described by a common framework that consists of a message, update, and readout function. For each node, the message function generates a representation, i.e., message by aggregating the features of the node, its neighbor nodes, and the associated edges. The update function updates the node representation using the message iteratively. The readout function generates a representation for the whole graph by aggregating all the node representations. Many variants of GNN models have been proposed in terms of the aggregation method for the message function. The convolution mechanism of CNN was used as an aggregation method for generating the node representation~\cite{duvenaud2015convolutional, kipf2017semi}. GraphSage~\cite{hamilton2017inductive} uniformly sampled a fixed-size set of neighbors within $k$ hops to keep the computational cost fixed and examined three different aggregation methods called mean aggregator, LSTM aggregator, and pooling aggregator. Attention-based aggregation was also proposed \cite{velivckovic2018graph}. For the update function, the iterative update function was proposed by early GNNs~\cite{gori2005new, scarselli2008graph}, and GRU was employed for such mechanism by Gated Graph Neural Network (GG-NN)~\cite{li2016gated}. Graph Isomorphism Network (GIN)~\cite{xu2019powerful} used the simple message and update function to capture the graph structure and showed good performance in graph classification tasks. Recently, there have been many efforts to incorporate Transformer in GNNs. Most of them take the encoder part of Transformer to use the power of its multi-head self-attention mechanism in aggregations and updates \cite{maziarka2020molecule, choi2020learning, koncel2019text, wang2020amr, xu2019multi, yun2019graph}.



Although such variants of GNNs have been applied to a variety of domains, the applications can be grouped into two categories: 1) classification or regression tasks, and 2) graph generation tasks. The tasks of the first category contain 
document classification based on citation networks \cite{kipf2017semi, hamilton2017inductive},
traffic forecast \cite{cui2019traffic,li2018diffusion},
link prediction in recommendation systems \cite{vdberg2017graph,ying2018graph},
learning dynamics of physical systems \cite{battaglia2016interaction},
molecular properties and protein interfaces prediction \cite{gilmer2017neural,klicpera2020directional, fout2017protein},
cancer type classification and polypharmacy's side-effect prediction in biomedical domains \cite{rhee2018hybrid,zitnik2018modeling},
reasoning about knowledge graphs \cite{hamaguchi2017knowledge,wang2018cross},
classification and segmentation of images, videos and 3D meshes \cite{wang2018non,wang2019dynamic},
classification of regions in images \cite{chen2018iterative},
text generation \cite{koncel2019text},
and sketch recognition \cite{xu2019multi}.

For the graph generation tasks, graphs can be generated in two different approaches: 1) non-auto-regressive approach and 2) auto-regressive approach. As an non-auto-regressive approach, \citet{grover2018graphite} proposed an iterative decoding method over a single fixed set of nodes based on a GNN-based encoder. In addition, variational autoencoder (VAE) has been used for the various types of graph generation~\cite{kipf2016variational, ma2018constrained, simonovsky2018graphvae, jin2018junction} and has shown meaningful results, especially for molecule generation tasks, where detailed attributes of nodes and edges are required to be generated as well. Such non-auto-regressive approach, however, has not shown effectiveness for large-scale graph generation. As an auto-regressive approach, GraphRNN model~\cite{you2018graphrnn} generates new nodes through the graph-level RNN while generating the edges for each newly generated nodes through the edge-level RNN. Moreover, Graph Recurrent Attention Network (GRAN)~\cite{liao2019efficient} enables the block-wise generation of nodes and associated edges where the block size is controllable, which enables large-scale graph generation. However, GRAN has not shown its capability of predicting possible detailed attributes associated with nodes and edges (such as node labels, edge types, etc.) other than just predicting the existence of the nodes and edges.

GRAT is the general-purpose, graph-to-graph model that can be applied to not only generation but also classification or regression of graph data by enabling the self-attention mechanism to be aware of edge information. To the best of our knowledge, GRAT is the first Transformer-based general-purpose, graph-to-graph model. \citet{jin2019hierarchical} is closely related to our work in terms of the graph-to-graph translation model in auto-regressive manner. However, their coverage of applications can be limited since their generation is nothing but growing the given input graph through subsequent attachments. Lastly, Graph2Graph Transformer~\cite{mohammadshahi2019graph} only uses the encoder part of Transformer, i.e., the multi-head self-attention is not fully utilized in its own decoder called Transition Classifier. Also, it may not generate the entire graph, instead, it can only generate an edge for a given two nodes.

\section{Conclusion} 
In this work, we presented Graph-Aware Transformer (GRAT), a general and flexible encoder-decoder model for graph applications. By fully utilizing the power of multi-head self-attention from Transformer, GRAT outperformed the existing state-of-the-art models in the graph property prediction tasks. Also, GRAT applied the graph-to-graph translation approach to the task of predicting resulting products in organic chemical reactions for the first time and showed the competitive performance over existing methods. In the future, we plan to apply GRAT to tasks in other domains rather than the chemistry, such as physics or knowledge graphs, to prove its generality. Alongside that, we also would like to find a way to incorporate domain-specific (i.e., chemistry or materials science) knowledge into the model without detriment to the general Transformer architecture.

\small

\bibliographystyle{unsrtnat}
\bibliography{neurips_2020}

\begin{thebibliography}{59}
\providecommand{\natexlab}[1]{#1}
\providecommand{\url}[1]{\texttt{#1}}
\expandafter\ifx\csname urlstyle\endcsname\relax
  \providecommand{\doi}[1]{doi: #1}\else
  \providecommand{\doi}{doi: \begingroup \urlstyle{rm}\Url}\fi

\bibitem[Zhou et~al.(2018)Zhou, Cui, Zhang, Yang, Liu, and Sun]{Zhou2018corr}
Jie Zhou, Ganqu Cui, Zhengyan Zhang, Cheng Yang, Zhiyuan Liu, and Maosong Sun.
\newblock Graph neural networks: {A} review of methods and applications.
\newblock \emph{CoRR}, abs/1812.08434, 2018.
\newblock URL \url{http://arxiv.org/abs/1812.08434}.

\bibitem[Zhang et~al.(2019)Zhang, He, and Katabi]{pmlr-v97-zhang19e}
Guo Zhang, Hao He, and Dina Katabi.
\newblock Circuit-{GNN}: Graph neural networks for distributed circuit design.
\newblock In Kamalika Chaudhuri and Ruslan Salakhutdinov, editors,
  \emph{Proceedings of the 36th International Conference on Machine Learning},
  volume~97 of \emph{Proceedings of Machine Learning Research}, pages
  7364--7373, Long Beach, California, USA, 09--15 Jun 2019. PMLR.
\newblock URL \url{http://proceedings.mlr.press/v97/zhang19e.html}.

\bibitem[Hamaguchi et~al.(2017)Hamaguchi, Oiwa, Shimbo, and
  Matsumoto]{hamaguchi2017knowledge}
Takuo Hamaguchi, Hidekazu Oiwa, Masashi Shimbo, and Yuji Matsumoto.
\newblock Knowledge transfer for out-of-knowledge-base entities: A graph neural
  network approach.
\newblock In \emph{Proceedings of the Twenty-Sixth International Joint
  Conference on Artificial Intelligence (IJCAI-17)}, 2017.

\bibitem[Fan et~al.(2019)Fan, Ma, Li, He, Zhao, Tang, and Yin]{wenqi2019www}
Wenqi Fan, Yao Ma, Qing Li, Yuan He, Eric Zhao, Jiliang Tang, and Dawei Yin.
\newblock Graph neural networks for social recommendation.
\newblock In \emph{The World Wide Web Conference}, WWW ’19, page 417–426,
  New York, NY, USA, 2019. Association for Computing Machinery.
\newblock ISBN 9781450366748.
\newblock \doi{10.1145/3308558.3313488}.
\newblock URL \url{https://doi.org/10.1145/3308558.3313488}.

\bibitem[Butler et~al.(2018)Butler, Davies, Cartwright, Isayev, and
  Walsh]{butler2018ml}
Keith~T. Butler, Daniel~W. Davies, Hugh Cartwright, Olexandr Isayev, and Aron
  Walsh.
\newblock Machine learning for molecular and materials science.
\newblock \emph{Nature}, 559:\penalty0 547--555, 2018.

\bibitem[Vaswani et~al.(2017)Vaswani, Shazeer, Parmar, Uszkoreit, Jones, Gomez,
  Kaiser, and Polosukhin]{vaswani2017}
A.~Vaswani, N.~Shazeer, N.~Parmar, J.~Uszkoreit, L.~Jones, A.N. Gomez,
  L.~Kaiser, and I.~Polosukhin.
\newblock Attention is all you need.
\newblock In \emph{Advances in Neural Information Processing Systems 31}, 2017.

\bibitem[Maziarka et~al.(2020)Maziarka, Danel, Mucha, Rataj, Tabor, and
  Jastrz{\k{e}}bski]{maziarka2020molecule}
{\L}ukasz Maziarka, Tomasz Danel, S{\l}awomir Mucha, Krzysztof Rataj, Jacek
  Tabor, and Stanis{\l}aw Jastrz{\k{e}}bski.
\newblock Molecule attention transformer.
\newblock \emph{arXiv preprint arXiv:2002.08264}, 2020.

\bibitem[Dumoulin et~al.(2018)Dumoulin, Perez, Schucher, Strub, Vries,
  Courville, and Bengio]{dumoulin2018}
V.~Dumoulin, E.~Perez, N.~Schucher, F.~Strub, H.d. Vries, A.~Courville, and
  Y.~Bengio.
\newblock Feature-wise transformations.
\newblock \emph{Distill}, 2018.
\newblock \doi{10.23915/distill.00011}.
\newblock https://distill.pub/2018/feature-wise-transformations.

\bibitem[Devlin et~al.(2019)Devlin, Chang, Lee, and Toutanova]{devlin2019naacl}
Jacob Devlin, Ming-Wei Chang, Kenton Lee, and Kristina Toutanova.
\newblock {BERT}: Pre-training of deep bidirectional transformers for language
  understanding.
\newblock In \emph{Proceedings of the 2019 Conference of the North {A}merican
  Chapter of the Association for Computational Linguistics: Human Language
  Technologies, Volume 1 (Long and Short Papers)}, pages 4171--4186,
  Minneapolis, Minnesota, June 2019. Association for Computational Linguistics.
\newblock \doi{10.18653/v1/N19-1423}.
\newblock URL \url{https://www.aclweb.org/anthology/N19-1423}.

\bibitem[Hu et~al.(2020)Hu, Liu, Gomes, Zitnik, Liang, Pande, and
  Leskovec]{hu2020pretrain}
Weihua Hu, Bowen Liu, Joseph Gomes, Marinka Zitnik, Percy Liang, Vijay~S.
  Pande, and Jure Leskovec.
\newblock Strategies for pre-training graph neural networks.
\newblock In \emph{Proceedings of the International Conference on Learning
  Representations (ICLR)}, 2020.

\bibitem[Wu et~al.(2018)Wu, Ramsundar, Feinberg, Gomes, Geniesse, Pappu,
  Leswing, and Pande]{zhenqin2018rsc}
Zhenqin Wu, Bharath Ramsundar, Evan N. Feinberg, Joseph Gomes, Caleb Geniesse,
  Aneesh~S. Pappu, Karl Leswing, and Vijay Pande.
\newblock Moleculenet: a benchmark for molecular machine learning.
\newblock \emph{Chem. Sci.}, 9:\penalty0 513--530, 2018.
\newblock \doi{10.1039/C7SC02664A}.
\newblock URL \url{http://dx.doi.org/10.1039/C7SC02664A}.

\bibitem[Gilmer et~al.(2017)Gilmer, Schoenholz, Riley, Vinyals, and
  Dahl]{gilmer2017neural}
Justin Gilmer, Samuel~S Schoenholz, Patrick~F Riley, Oriol Vinyals, and
  George~E Dahl.
\newblock Neural message passing for quantum chemistry.
\newblock In \emph{Proceedings of the 34th International Conference on Machine
  Learning}, pages 1263--1272. JMLR. org, 2017.

\bibitem[Klicpera et~al.(2020)Klicpera, Gro{\ss}, and
  G{\"u}nnemann]{klicpera2020directional}
Johannes Klicpera, Janek Gro{\ss}, and Stephan G{\"u}nnemann.
\newblock Directional message passing for molecular graphs.
\newblock In \emph{Proceedings of the International Conference on Learning
  Representations (ICLR)}, 2020.

\bibitem[Ruddigkeit et~al.(2012)Ruddigkeit, van Deursen, Blum, and
  Reymond]{Ruddigkeit2012}
Lars Ruddigkeit, Ruud van Deursen, Lorenz~C. Blum, and Jean-Louis Reymond.
\newblock Enumeration of 166 billion organic small molecules in the chemical
  universe database gdb-17.
\newblock \emph{Journal of Chemical Information and Modeling}, 52\penalty0
  (11):\penalty0 2864--2875, 2012.
\newblock \doi{10.1021/ci300415d}.
\newblock URL \url{https://doi.org/10.1021/ci300415d}.
\newblock PMID: 23088335.

\bibitem[Landrum()]{rdkit}
Greg Landrum.
\newblock Rdkit: Open-source cheminformatics.
\newblock URL \url{http://www.rdkit.org}.

\bibitem[Lowe(2014)]{lowe2014}
D.~M. Lowe.
\newblock Patent reaction extraction: downloads;
  https://bitbucket.org/dan2097/patent-reaction-extraction/downloads.
\newblock 2014.

\bibitem[Jin et~al.(2017)Jin, Coley, Barzilay, and Jaakkola]{jin2017}
W.~Jin, C.~Coley, R.~Barzilay, and T.~Jaakkola.
\newblock Predicting organic reaction outcomes with weisfeiler-lehman network.
\newblock In \emph{Advances in Neural Information Processing Systems 31}, 2017.

\bibitem[Do et~al.(2019)Do, Tran, and Venkatesh]{do2019}
K.~Do, T.~Tran, and S.~Venkatesh.
\newblock Graph transformation policy network for chemical reaction prediction.
\newblock In \emph{Proceedings of the 25th ACM SigKDD International Conference
  on Knowledge Discovery \& Data Mining}, pages 750--760, 2019.

\bibitem[Coley et~al.(2019)Coley, Jin, Rogers, Jamison, Jaakkola, Green,
  Barzilay, and Jensen]{coley2019}
C.W. Coley, W.~Jin, L.~Rogers, T.F. Jamison, T.S. Jaakkola, W.H. Green,
  R.~Barzilay, and K.F. Jensen.
\newblock A graph-convolutional neural network model for the prediction of
  chemical reactivity.
\newblock \emph{Chemical Science}, 10:\penalty0 370--377, 2019.

\bibitem[Schwaller et~al.(2019)Schwaller, Laino, Gaudin, Bolgar, Hunter, Bekas,
  and Lee]{schwaller2019}
P.~Schwaller, T.~Laino, T.~Gaudin, P.~Bolgar, C.A. Hunter, C.~Bekas, and A.A.
  Lee.
\newblock Molecular transformer: A model for uncertainty-calibrated chemical
  reaction prediction.
\newblock \emph{ACS Central Science}, 5:\penalty0 1572--1583, 2019.

\bibitem[Schwaller et~al.(2018)Schwaller, Gaudin, Lanyi, Bekas, and
  Laino]{schwaller2018}
P.~Schwaller, T.~Gaudin, D.~Lanyi, C.~Bekas, and T.~Laino.
\newblock Found in translation: Predicting outcomes of complex organic
  chemistry reactions using neural sequence-to-sequence models.
\newblock \emph{Chemical Science}, 9:\penalty0 6091--6098, 2018.

\bibitem[Duvenaud et~al.(2015)Duvenaud, Maclaurin, Iparraguirre, Bombarell,
  Hirzel, Aspuru-Guzik, and Adams]{duvenaud2015convolutional}
David~K Duvenaud, Dougal Maclaurin, Jorge Iparraguirre, Rafael Bombarell,
  Timothy Hirzel, Al{\'a}n Aspuru-Guzik, and Ryan~P Adams.
\newblock Convolutional networks on graphs for learning molecular fingerprints.
\newblock In \emph{Advances in neural information processing systems}, pages
  2224--2232, 2015.

\bibitem[Kipf and Welling(2017)]{kipf2017semi}
Thomas~N Kipf and Max Welling.
\newblock Semi-supervised classification with graph convolutional networks.
\newblock In \emph{Proceedings of the International Conference on Learning
  Representations (ICLR)}, 2017.

\bibitem[Hamilton et~al.(2017)Hamilton, Ying, and
  Leskovec]{hamilton2017inductive}
Will Hamilton, Zhitao Ying, and Jure Leskovec.
\newblock Inductive representation learning on large graphs.
\newblock In \emph{Advances in neural information processing systems}, pages
  1024--1034, 2017.

\bibitem[Veli{\v{c}}kovi{\'c} et~al.(2018)Veli{\v{c}}kovi{\'c}, Cucurull,
  Casanova, Romero, Lio, and Bengio]{velivckovic2018graph}
Petar Veli{\v{c}}kovi{\'c}, Guillem Cucurull, Arantxa Casanova, Adriana Romero,
  Pietro Lio, and Yoshua Bengio.
\newblock Graph attention networks.
\newblock In \emph{Proceedings of the International Conference on Learning
  Representations (ICLR)}, 2018.

\bibitem[Gori et~al.(2005)Gori, Monfardini, and Scarselli]{gori2005new}
Marco Gori, Gabriele Monfardini, and Franco Scarselli.
\newblock A new model for learning in graph domains.
\newblock In \emph{Proceedings. 2005 IEEE International Joint Conference on
  Neural Networks, 2005.}, volume~2, pages 729--734. IEEE, 2005.

\bibitem[Scarselli et~al.(2008)Scarselli, Gori, Tsoi, Hagenbuchner, and
  Monfardini]{scarselli2008graph}
Franco Scarselli, Marco Gori, Ah~Chung Tsoi, Markus Hagenbuchner, and Gabriele
  Monfardini.
\newblock The graph neural network model.
\newblock \emph{IEEE Transactions on Neural Networks}, 20\penalty0
  (1):\penalty0 61--80, 2008.

\bibitem[Li et~al.(2016)Li, Zemel, Brockschmidt, and Tarlow]{li2016gated}
Yujia Li, Richard Zemel, Marc Brockschmidt, and Daniel Tarlow.
\newblock Gated graph sequence neural networks.
\newblock In \emph{Proceedings of ICLR'16}, April 2016.
\newblock URL
  \url{https://www.microsoft.com/en-us/research/publication/gated-graph-sequence-neural-networks/}.

\bibitem[Xu et~al.(2019{\natexlab{a}})Xu, Hu, Leskovec, and
  Jegelka]{xu2019powerful}
Keyulu Xu, Weihua Hu, Jure Leskovec, and Stefanie Jegelka.
\newblock How powerful are graph neural networks?
\newblock In \emph{Proceedings of the International Conference on Learning
  Representations (ICLR)}, 2019{\natexlab{a}}.

\bibitem[Choi et~al.(2020)Choi, Xu, Li, Dusenberry, Flores, Xue, and
  Dai]{choi2020learning}
Edward Choi, Zhen Xu, Yujia Li, Michael~W Dusenberry, Gerardo Flores, Emily
  Xue, and Andrew~M Dai.
\newblock Learning the graphical structure of electronic health records with
  graph convolutional transformer.
\newblock In \emph{Association for the Advancement of Artificial Intelligence},
  2020.

\bibitem[Rik Koncel-Kedziorski and Hajishirzi(2019)]{koncel2019text}
Yi~Luan Mirella~Lapata Rik Koncel-Kedziorski, Dhanush~Bekal and Hannaneh
  Hajishirzi.
\newblock Text generation from knowledge graphs with graph transformers.
\newblock In \emph{NAACL}, 2019.

\bibitem[Wang et~al.(2020)Wang, Wan, and Jin]{wang2020amr}
Tianming Wang, Xiaojun Wan, and Hanqi Jin.
\newblock Amr-to-text generation with graph transformer.
\newblock \emph{Transactions of the Association for Computational Linguistics},
  8:\penalty0 19--33, 2020.

\bibitem[Xu et~al.(2019{\natexlab{b}})Xu, Joshi, and Bresson]{xu2019multi}
Peng Xu, Chaitanya~K Joshi, and Xavier Bresson.
\newblock Multi-graph transformer for free-hand sketch recognition.
\newblock \emph{arXiv preprint arXiv:1912.11258}, 2019{\natexlab{b}}.

\bibitem[Yun et~al.(2019)Yun, Jeong, Kim, Kang, and Kim]{yun2019graph}
Seongjun Yun, Minbyul Jeong, Raehyun Kim, Jaewoo Kang, and Hyunwoo~J Kim.
\newblock Graph transformer networks.
\newblock In \emph{Advances in Neural Information Processing Systems}, pages
  11960--11970, 2019.

\bibitem[Cui et~al.(2019)Cui, Henrickson, Ke, and Wang]{cui2019traffic}
Zhiyong Cui, Kristian Henrickson, Ruimin Ke, and Yinhai Wang.
\newblock Traffic graph convolutional recurrent neural network: A deep learning
  framework for network-scale traffic learning and forecasting.
\newblock \emph{IEEE Transactions on Intelligent Transportation Systems}, 2019.

\bibitem[Li et~al.(2018)Li, Yu, Shahabi, and Liu]{li2018diffusion}
Yaguang Li, Rose Yu, Cyrus Shahabi, and Yan Liu.
\newblock Diffusion convolutional recurrent neural network: Data-driven traffic
  forecasting.
\newblock In \emph{Proceedings of the International Conference on Learning
  Representations (ICLR)}, 2018.

\bibitem[van~den Berg et~al.(2017)van~den Berg, Kipf, and
  Welling]{vdberg2017graph}
Rianne van~den Berg, Thomas~N Kipf, and Max Welling.
\newblock Graph convolutional matrix completion.
\newblock \emph{arXiv preprint arXiv:1706.02263}, 2017.

\bibitem[Ying et~al.(2018)Ying, He, Chen, Eksombatchai, Hamilton, and
  Leskovec]{ying2018graph}
Rex Ying, Ruining He, Kaifeng Chen, Pong Eksombatchai, William~L Hamilton, and
  Jure Leskovec.
\newblock Graph convolutional neural networks for web-scale recommender
  systems.
\newblock In \emph{Proceedings of the 24th ACM SIGKDD International Conference
  on Knowledge Discovery \& Data Mining}, pages 974--983, 2018.

\bibitem[Battaglia et~al.(2016)Battaglia, Pascanu, Lai, Rezende,
  et~al.]{battaglia2016interaction}
Peter Battaglia, Razvan Pascanu, Matthew Lai, Danilo~Jimenez Rezende, et~al.
\newblock Interaction networks for learning about objects, relations and
  physics.
\newblock In \emph{Advances in neural information processing systems}, pages
  4502--4510, 2016.

\bibitem[Fout et~al.(2017)Fout, Byrd, Shariat, and Ben-Hur]{fout2017protein}
Alex Fout, Jonathon Byrd, Basir Shariat, and Asa Ben-Hur.
\newblock Protein interface prediction using graph convolutional networks.
\newblock In \emph{Advances in neural information processing systems}, pages
  6530--6539, 2017.

\bibitem[Rhee et~al.(2018)Rhee, Seo, and Kim]{rhee2018hybrid}
Sungmin Rhee, Seokjun Seo, and Sun Kim.
\newblock Hybrid approach of relation network and localized graph convolutional
  filtering for breast cancer subtype classification.
\newblock In \emph{Proceedings of the Twenty-Seventh International Joint
  Conference on Artificial Intelligence (IJCAI-18)}, 2018.

\bibitem[Zitnik et~al.(2018)Zitnik, Agrawal, and Leskovec]{zitnik2018modeling}
Marinka Zitnik, Monica Agrawal, and Jure Leskovec.
\newblock Modeling polypharmacy side effects with graph convolutional networks.
\newblock \emph{Bioinformatics}, 34\penalty0 (13):\penalty0 i457--i466, 2018.

\bibitem[Wang et~al.(2018{\natexlab{a}})Wang, Lv, Lan, and
  Zhang]{wang2018cross}
Zhichun Wang, Qingsong Lv, Xiaohan Lan, and Yu~Zhang.
\newblock Cross-lingual knowledge graph alignment via graph convolutional
  networks.
\newblock In \emph{Proceedings of the 2018 Conference on Empirical Methods in
  Natural Language Processing}, pages 349--357, 2018{\natexlab{a}}.

\bibitem[Wang et~al.(2018{\natexlab{b}})Wang, Girshick, Gupta, and
  He]{wang2018non}
Xiaolong Wang, Ross Girshick, Abhinav Gupta, and Kaiming He.
\newblock Non-local neural networks.
\newblock In \emph{Proceedings of the IEEE conference on computer vision and
  pattern recognition}, pages 7794--7803, 2018{\natexlab{b}}.

\bibitem[Wang et~al.(2019)Wang, Sun, Liu, Sarma, Bronstein, and
  Solomon]{wang2019dynamic}
Yue Wang, Yongbin Sun, Ziwei Liu, Sanjay~E Sarma, Michael~M Bronstein, and
  Justin~M Solomon.
\newblock Dynamic graph cnn for learning on point clouds.
\newblock \emph{ACM Transactions on Graphics (TOG)}, 38\penalty0 (5):\penalty0
  1--12, 2019.

\bibitem[Chen et~al.(2018)Chen, Li, Fei-Fei, and Gupta]{chen2018iterative}
Xinlei Chen, Li-Jia Li, Li~Fei-Fei, and Abhinav Gupta.
\newblock Iterative visual reasoning beyond convolutions.
\newblock In \emph{Proceedings of the IEEE Conference on Computer Vision and
  Pattern Recognition}, pages 7239--7248, 2018.

\bibitem[Grover et~al.(2018)Grover, Zweig, and Ermon]{grover2018graphite}
Aditya Grover, Aaron Zweig, and Stefano Ermon.
\newblock Graphite: Iterative generative modeling of graphs.
\newblock \emph{arXiv preprint arXiv:1803.10459}, 2018.

\bibitem[Kipf and Welling(2016)]{kipf2016variational}
Thomas~N Kipf and Max Welling.
\newblock Variational graph auto-encoders.
\newblock \emph{arXiv preprint arXiv:1611.07308}, 2016.

\bibitem[Ma et~al.(2018)Ma, Chen, and Xiao]{ma2018constrained}
Tengfei Ma, Jie Chen, and Cao Xiao.
\newblock Constrained generation of semantically valid graphs via regularizing
  variational autoencoders.
\newblock In \emph{Advances in Neural Information Processing Systems}, pages
  7113--7124, 2018.

\bibitem[Simonovsky and Komodakis(2018)]{simonovsky2018graphvae}
Martin Simonovsky and Nikos Komodakis.
\newblock Graphvae: Towards generation of small graphs using variational
  autoencoders.
\newblock In \emph{International Conference on Artificial Neural Networks},
  pages 412--422. Springer, 2018.

\bibitem[Jin et~al.(2018)Jin, Barzilay, and Jaakkola]{jin2018junction}
Wengong Jin, Regina Barzilay, and Tommi Jaakkola.
\newblock Junction tree variational autoencoder for molecular graph generation.
\newblock \emph{arXiv preprint arXiv:1802.04364}, 2018.

\bibitem[You et~al.(2018)You, Ying, Ren, Hamilton, and
  Leskovec]{you2018graphrnn}
Jiaxuan You, Rex Ying, Xiang Ren, William~L Hamilton, and Jure Leskovec.
\newblock Graphrnn: Generating realistic graphs with deep auto-regressive
  models.
\newblock \emph{arXiv preprint arXiv:1802.08773}, 2018.

\bibitem[Liao et~al.(2019)Liao, Li, Song, Wang, Hamilton, Duvenaud, Urtasun,
  and Zemel]{liao2019efficient}
Renjie Liao, Yujia Li, Yang Song, Shenlong Wang, Will Hamilton, David~K
  Duvenaud, Raquel Urtasun, and Richard Zemel.
\newblock Efficient graph generation with graph recurrent attention networks.
\newblock In \emph{Advances in Neural Information Processing Systems}, pages
  4257--4267, 2019.

\bibitem[Jin et~al.(2019)Jin, Barzilay, and Jaakkola]{jin2019hierarchical}
W~Jin, R~Barzilay, and T~Jaakkola.
\newblock Hierarchical graph-to-graph translation for molecules.
\newblock \emph{arXiv preprint arXiv:1907.11223}, 2019.

\bibitem[Mohammadshahi and Henderson(2019)]{mohammadshahi2019graph}
Alireza Mohammadshahi and James Henderson.
\newblock Graph-to-graph transformer for transition-based dependency parsing.
\newblock \emph{arXiv preprint arXiv:1911.03561}, 2019.

\bibitem[Abadi et~al.(2015)Abadi, Agarwal, Barham, Brevdo, Chen, Citro,
  Corrado, Davis, Dean, Devin, Ghemawat, Goodfellow, Harp, Irving, Isard, Jia,
  Jozefowicz, Kaiser, Kudlur, Levenberg, Man\'{e}, Monga, Moore, Murray, Olah,
  Schuster, Shlens, Steiner, Sutskever, Talwar, Tucker, Vanhoucke, Vasudevan,
  Vi\'{e}gas, Vinyals, Warden, Wattenberg, Wicke, Yu, and
  Zheng]{tensorflow2015-whitepaper}
Mart\'{\i}n Abadi, Ashish Agarwal, Paul Barham, Eugene Brevdo, Zhifeng Chen,
  Craig Citro, Greg~S. Corrado, Andy Davis, Jeffrey Dean, Matthieu Devin,
  Sanjay Ghemawat, Ian Goodfellow, Andrew Harp, Geoffrey Irving, Michael Isard,
  Yangqing Jia, Rafal Jozefowicz, Lukasz Kaiser, Manjunath Kudlur, Josh
  Levenberg, Dandelion Man\'{e}, Rajat Monga, Sherry Moore, Derek Murray, Chris
  Olah, Mike Schuster, Jonathon Shlens, Benoit Steiner, Ilya Sutskever, Kunal
  Talwar, Paul Tucker, Vincent Vanhoucke, Vijay Vasudevan, Fernanda Vi\'{e}gas,
  Oriol Vinyals, Pete Warden, Martin Wattenberg, Martin Wicke, Yuan Yu, and
  Xiaoqiang Zheng.
\newblock {TensorFlow}: Large-scale machine learning on heterogeneous systems,
  2015.
\newblock URL \url{https://www.tensorflow.org/}.
\newblock Software available from tensorflow.org.

\bibitem[Kuchaiev et~al.(2018)Kuchaiev, Ginsburg, Gitman, Lavrukhin, Li,
  Nguyen, Case, and Micikevicius]{openseq2seq2018}
Oleksii Kuchaiev, Boris Ginsburg, Igor Gitman, Vitaly Lavrukhin, Jason Li,
  Huyen Nguyen, Carl Case, and Paulius Micikevicius.
\newblock Mixed-precision training for nlp and speech recognition with
  openseq2seq.
\newblock \emph{arXiv preprint arXiv:1805.10387}, 2018.

\bibitem[Kingma and Ba(2014)]{kingma2014adam}
Diederik Kingma and Jimmy Ba.
\newblock Adam: A method for stochastic optimization.
\newblock \emph{arXiv preprint arXiv:1412.6980}, 2014.

\bibitem[Daylight Chemical Information~Systems()]{smiles}
Inc. Daylight Chemical Information~Systems.
\newblock Smiles - a simplified chemical language.
\newblock URL
  \url{https://www.daylight.com/dayhtml/doc/theory/theory.smiles.html}.

\end{thebibliography}

\appendix
\section*{Appendix}
\renewcommand{\thesubsection}{\Alph{subsection}}
\subsection{Additional Details for Section~\ref{section:property_prediction}}
\label{section:add_detail_4.1}
Our model has 32 self-attention layers with 32 heads, node representations of size 256, and feed-forward layers with inner dimension 1024. As described in Section~\ref{section:pretraining_strategies}, the graph-level embedding resulted from the <CLS> token is fed into a fully connected layer with the hidden size 512 which generates 12 outputs corresponding to the properties of the 12 regression tasks. $f_a$ consists of hidden states of size 32 with $tanh$ activation function and final outputs of size 64 ($\gamma$ and $\beta$ for each layer).

The code is implemented in Tensorflow \cite{tensorflow2015-whitepaper}, and the self-attention is based on NVIDIA OpenSeq2Seq \cite{openseq2seq2018}. We used one GPU (Nvidia v100). The batch size is 50, and all models were trained using SGD with the ADAM optimizer \cite{kingma2014adam}.

We add the atom group number, the atom weight, and the formal charge for each atom as an additional input node/atom features. Note that those are commonly used atom features in the chemical domain tasks, and are easily obtained by RDKit \cite{rdkit}. 

In addtion, the 11 molecule descriptors (obtained by RDKit) used for the pre-training are listed below:

\begin{itemize}
\item Number of atoms
\item Topological polar surface area (TPSA)
\item logP
\item Molecular mass
\item Molecular weight
\item Number of valence electrons
\item Number of aromatic rings
\item Number of saturated rings
\item Number of aliphatic rings
\item Balaban's J index (BalabanJ)
\item Bertz CT (BertzCT)
\end{itemize}

\subsection{Additional Details for Section~\ref{section:chemical_reaction_outcomes_prediction}}
Since chemical compounds in the USPTO dataset are represented in Simplified Molecular Input Line Entry System (SMILES) \cite{smiles}, which is a sequence of characters designed to describe the chemical structure, we convert the input compounds to input graph structures using RDKit \cite{rdkit} and convert the output graph structures back to SMILES once finishing the prediction to compare the predicted graph with the ground truth in SMILES.

Since the tokens in GRAT do not need to contain the edge information, the vocabulary size is the same as the number of elements in the periodic table plus a few special tokens such as <G> and <EOG>. This makes our model enable to maintain small parameters (about 8M) despite of many layers. (The model proposed by \citet{schwaller2019} has 12M parameters.) We add an additional special token, one of <REACTANT>, <REAGENT>, and <PRODUCT>, at the start of each graph as a delimiter to explicitly distinguish each graph from others as is done in~\cite{schwaller2019}. These special tokens are also connected to all other atom nodes by introducing an additional virtual edge type. Since SMILES imposes a canonical order on participating atoms in the compound, we impose the same order on the atoms/nodes in the corresponding graph through the positional encoding for the graph translation task.


Both of the encoder and the decoder have 24 self-attention layers with 8 heads, node representations of size 128, and feed-forward layers with inner dimension 256. $f_a$ for the encoder and the decoder each consists of hidden states of size 32 with $tanh$ activation function and final outputs of size 48 ($\gamma$ and $\beta$ for each layer). We assume that there is a "no bond" edge type for the encoder because we want to utilize the distance information between every pair of nodes. However, for the decoder, we do not use such "no bond" edge by making attention values corresponding to "no bond" zero so that two disconnected nodes cannot attend on each other as described in Section~\ref{section:graph_encoding}.

As in Section~\ref{section:add_detail_4.1}, the code is implemented in Tensorflow and based on NVIDIA OpenSeq2Seq. We used 8 GPUs (Nvidia p40), and the total batch size is 128. All models were trained using SGD with the ADAM optimizer \cite{kingma2014adam}. During decoding, the beam size is set to 8.

\end{document}